\documentclass[dvipsnames]{convergence}

\usepackage{wrapfig}
\usepackage{amsmath}
\usepackage{amssymb}
\usepackage{mathtools}
\usepackage{amsthm}

\theoremstyle{plain}

\theoremstyle{definition}

\theoremstyle{remark}

\usepackage[textsize=tiny]{todonotes}

\usepackage{lipsum}
\usepackage{bm}
\usepackage{multirow}
\usepackage{wrapfig}
\usepackage{xcolor}
\usepackage{svg}
\usepackage{subcaption}

\newcommand{\fullname}{Large Memory Model}
\newcommand{\name}{LM2}
\definecolor{gray}{HTML}{999999}
\definecolor{pink}{HTML}{F19C99}
\definecolor{blue}{HTML}{DAE8FC}
\definecolor{green}{HTML}{B5C4B1}
\definecolor{rose}{HTML}{EBD0D2}
\usepackage{microtype}
\usepackage{graphicx}
\usepackage{booktabs} % for professional tables

\usepackage{colortbl}

\usepackage{amsmath}
\usepackage{amssymb}
\usepackage{mathtools}
\usepackage{amsthm}

\usepackage{colortbl}
\usepackage{nicematrix}
\usepackage{amsthm}
\usepackage{amsmath,amsfonts,bm}
\usepackage{nicefrac}
\usepackage{array}
\usepackage{url}
\usepackage{pbox}

\definecolor{bluex}{rgb}{0.27, 0.42, 0.81}
\definecolor{purplex}{HTML}{9564bf}
\definecolor{red3}{HTML}{C52A20}
\definecolor{red2}{HTML}{B36A6F}
\definecolor{red1}{HTML}{FFb5b5}
\definecolor{purple}{HTML}{B36A6F}
\definecolor{darkyellow}{HTML}{D5BA82}
\definecolor{blue1}{HTML}{508AB2}
\definecolor{blue2}{HTML}{C4E4E3}
\definecolor{green1}{HTML}{A1D0C7}
\definecolor{green2}{HTML}{BFF6BA}
\definecolor{green3}{HTML}{028100}
\definecolor{teal}{HTML}{508AB2}
\definecolor{purple1}{HTML}{8d3a94}
\definecolor{grey1}{HTML}{999999}

\usepackage{pifont}% http://ctan.org/pkg/pifont

\tcbuselibrary{most,listings,theorems}
\newtcbtheorem[number within=section]{explain}{Explanation}%
{colback=green2!5,colframe=blue1,fonttitle=\bfseries, left=.02in, right=.02in,bottom=.02in, top=.02in}{explain}

\title{LM2: Large Memory Models}
\runningtitle{LM2: Large Memory Models}

\author[1]{Jikun Kang}
\author[1]{Wenqi Wu}
\author[1]{Filippos Christianos}
\author[1]{Alex J. Chan}
\author[1]{Fraser Greenlee}
\author[1]{\newline George Thomas}
\author[1]{Marvin Purtorab}
\author[1]{and Andy Toulis.}
\affiliation[1]{Convergence Labs Ltd.}

\abstract{
This paper introduces the \emph{\fullname} (\name), a decoder-only Transformer architecture enhanced with an auxiliary memory module that aims to address the limitations of standard Transformers in multi-step reasoning, relational argumentation, and synthesizing information distributed over long contexts. 
The proposed \name\ incorporates a memory module that acts as a contextual representation repository, interacting with input tokens via cross attention and updating through gating mechanisms. 
To preserve the Transformer’s general-purpose capabilities, \name\ maintains the original information flow while integrating a complementary memory pathway.
Experimental results on the BABILong benchmark demonstrate that the \name model outperforms both the memory-augmented RMT model by 37.1\% and the baseline Llama-3.2 model by 86.3\% on average across tasks. 
\name\ exhibits exceptional capabilities in multi-hop inference, numerical reasoning, and large-context question-answering. 
On the MMLU dataset, it achieves a 5.0\% improvement over a pre-trained vanilla model, demonstrating that its memory module does not degrade performance on general tasks.
Further, in our analysis, we explore the memory interpretability, effectiveness of memory modules, and test-time behavior.
Our findings emphasize the importance of explicit memory in enhancing Transformer architectures.
}

\correspondence{Jikun Kang at \email{jikun@convergence.ai}}
\metadata[Website]{\url{https://convergence.ai}}
\metadata[Code]{\url{https://github.com/convergence-ai/lm2}}

\usepackage{graphicx} % Required for inserting images

\begin{document}

\maketitle

\section{Introduction}

Transformer-based models have achieved remarkable success. 
Landmark architectures such as GPT-3 \cite{brown2020language}, BERT \cite{kenton2019bert}, and Vision Transformers \cite{dosovitskiy2020image} have established state-of-the-art performance across a wide array of applications, including machine translation \cite{zhu2020incorporating}, text summarization \cite{liu-lapata-2019-hierarchical}, question-answering \cite{li2023laffi}, and image recognition \cite{dosovitskiy2020image}. 
As demonstrated by studies on large-scale models, their generalization capabilities improve significantly with increased data and model size, leading to emergent behaviors that extend beyond their original training objectives \cite{kaplan2020scaling, KangLYT0F24}.
Despite their significant contributions, current Transformer models encounter critical limitations when applied to long context reasoning tasks \cite{kuratov2024babilong}. 
For instance, in the \emph{needle-in-a-haystack} problem, models must answer questions that require reasoning across facts scattered throughout exceedingly long documents. 
Effectively addressing tasks with extensive context demands the model’s ability to discern essential information from vast amounts of irrelevant data. 

Recent memory-augmented architectures \citep[e.g.,][]{bulatov2022recurrentmemorytransformer, ko2024memreasoner} attempt to tackle these challenges by using recurrent prompts to track long context information. 
However, these architectures primarily summarize previous answers into prompts without fully integrating long-term information, leading to performance degradation over long contexts. 
For example, on Task 2 (see \cref{sec:babilong_dataset}), MemReasoner \cite{ko2024memreasoner} achieves a performance score of 60.6 for context lengths under 8K, but drops significantly to 18.5 when the context length exceeds 16K.
Additionally, these models are specifically tailored for memory-based tasks, thereby sacrificing the generalization capabilities inherent to large language models (LLMs).

\begin{wrapfigure}[30]{r}{0.5\textwidth}
    \centering
    \includegraphics[width=0.9\linewidth]{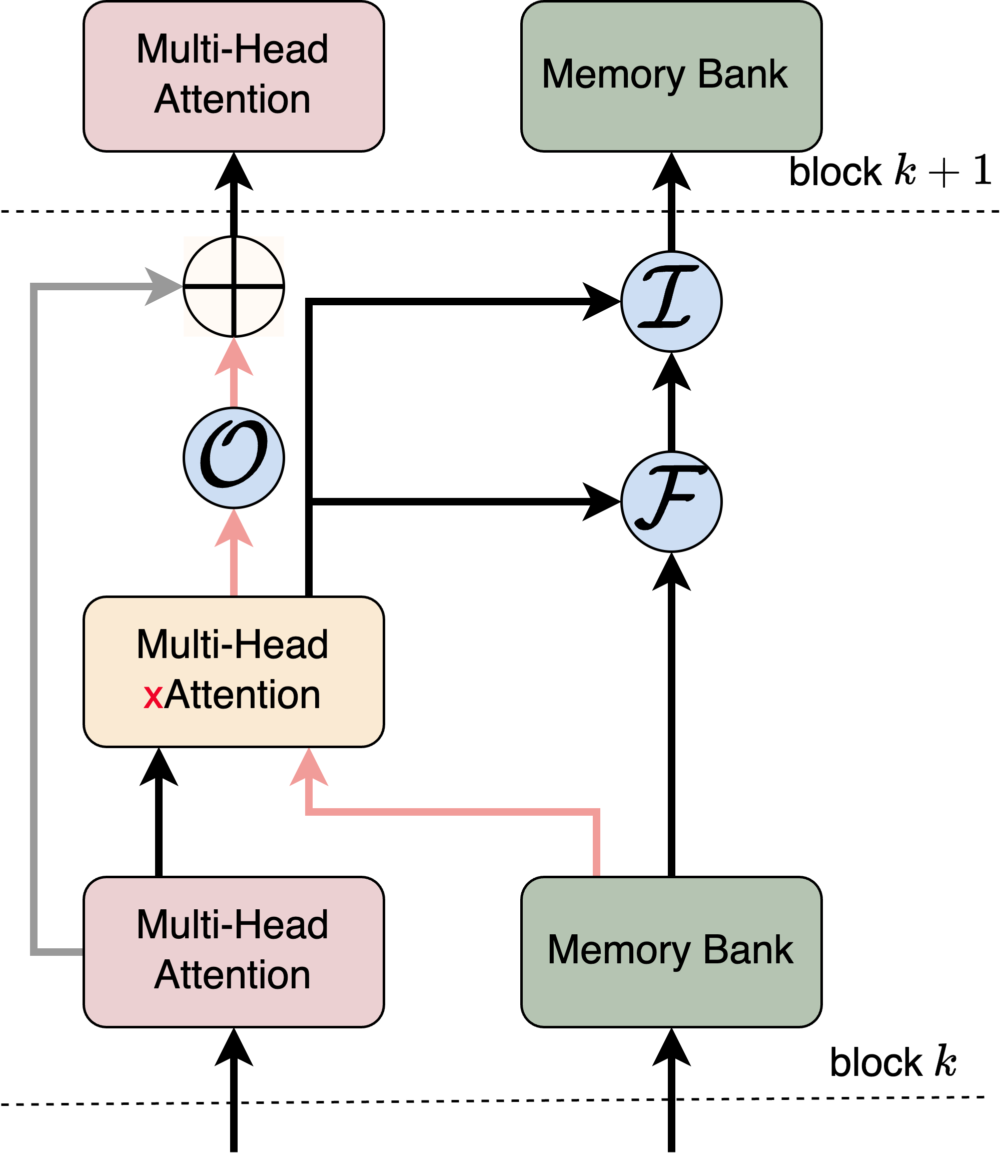}
    \caption{Illustration of \name\ overall architecture. It consists of a separate memory bank, which updates the main information flow through cross attention, and is updated using the input ($\mathcal{I}$), output ($\mathcal{O}$), and forget ($\mathcal{F}$) gates. For the information flow from one block to another, the \textcolor{gray}{gray} curve shows the normal attention flow and the \textcolor{pink}{pink} curve shows the extra memory flow.}
    \label{fig:lm2_wf}
\end{wrapfigure}

To address these limitations, we propose the \fullname~(\name), a novel architecture that enhances the Transformer framework with a dedicated memory module. This module functions as an auxiliary storage and retrieval mechanism, dynamically interacting with input embeddings to improve performance.
The memory module follows a structured process: initializing with a memory bank, leveraging cross attention for efficient interaction with sequence embeddings, and using gating mechanisms, such as forget and input gates, to selectively update stored information. 
By decoupling memory storage and retrieval from immediate processing, \name\ provides a robust solution for modeling long-term dependencies, overcoming the shortcomings of existing methods while maintaining computational efficiency. 
This architecture is particularly well-suited for tasks requiring long context and complex reasoning, offering a practical and scalable alternative to current approaches.

Moreover, as illustrated in Figure~\ref{fig:lm2_wf}, we maintain the original information flow—-namely, the output embeddings passed from one block to the next—-while introducing an additional, complementary memory information flow represented by the memory embeddings.
The memory information flow is controlled by a learnable output gate, which uses cross attention to dynamically regulate the amount of memory information passed to subsequent layers. 
This design ensures that the original attention information flow remains intact while dynamically incorporating relevant memory information as needed.
% By carefully regulating the memory flow with the output gate, the model maintains its capabilities on general LLM tasks without compromising the original information flow.

We first evaluate the effectiveness of \name\ on the \emph{BABILong} dataset \cite{kuratov2024babilong}, a challenging benchmark specifically designed to test memory-intensive reasoning capabilities. 
To verify that our memory-based approach does not undermine general performance, we also assess \name\ on the MMLU benchmark \cite{hendryckstest2021}, which spans a broad array of academic subjects and difficulty levels. 
Across both evaluations, \name\ outperforms state-of-the-art (SOTA) memory model Recurrent Memory Transformer (RMT) \cite{bulatov2022recurrentmemorytransformer} by up to 80.4\%, illustrating enhanced proficiency in multi-hop inference, numerical reasoning, and relational argumentation. These improvements underscore the value of incorporating our explicit memory mechanisms within Transformer architectures, enabling more robust handling of extended contexts. 

The contributions of this work are summarized as follows:
\begin{itemize}
    \item We propose a novel memory-augmented Transformer architecture that incorporates a dynamic memory module capable of capturing and leveraging long-term dependencies in sequential data.
    \item We introduce an additional memory information flow within the decoder block that complements the existing attention mechanism, enabling the integration of enriched memory information while preserving the original attention information.
    \item Through extensive experiments on long context reasoning tasks (up to a context of 128K tokens), \name\ outperforms SOTA memory-augmented model RMT and non-memory baseline Llama-3.2 on average 37.1\% and 86.3\%, respectively, demonstrating the practical benefits of our approach.
\end{itemize}

\section{\fullname~(\name)}

We present \fullname~ (\name), a memory-augmented Transformer model designed to enhance its long-term memory capabilities. 
\name\ consists of multiple Transformer decoder blocks, augmented with a memory module that dynamically stores and updates intermediate sequences of representations. 
The decoder block processes input sequences using positional embeddings, while the memory module interacts with these embeddings via cross attention mechanisms. 
We use a skip connection between the multi-head attention and the memory modules to facilitate learning and maintain the original intermediate embeddings of the Transformer.
The memory updates are controlled by learnable control gates, denoted as $\mathcal{F}$, $\mathcal{I}$, and $\mathcal{O}$, which correspond to the \emph{forget}, \emph{input}, and \emph{output} gates, respectively.
The memory module operates through two primary stages: memory information flow, and memory updates. 
Each of these stages is elaborated on in the following sections.

\subsection{Memory Information Flow}

% \paragraph{Memory Bank Initialization}
As depicted in Figure~\ref{fig:lm2_wf}, we introduce an explicit memory module, named the memory bank $\mathbf{M} \in \mathbb{R}^{N \times d \times d}$, designed to store long-term memory. 
Here, $N$ denotes the number of memory slots, while $d$ represents the hidden dimension of each slot. 
For simplicity, each memory slot is initialized as an identity matrix: $\mathbf{M}_r = \mathbf{I}_{d \times d}$, where $r \in \{1, \dots, N\}$ and $\mathbf{I}_{d \times d}$ is the identity matrix. 

We use a cross attention-based mechanism between the memory bank and input embeddings to locate memory slots that contain relevant information.
This approach is based on the idea that humans tend to store and group related information together (e.g., in Documentation Science and Archival Science \citep{DooleyArchivalAdvantage}).
Note that the input embeddings $\mathbf{E}$ are encoded by the positional encoder, which embeds the input tokens and persists the temporal correlations between states and actions. 
Concretely, each input embedding $\mathbf{E}$ acts as the \emph{query}, while the memory bank $\mathbf{M}$ serves as both the \emph{key} and the \emph{value} store. 
Intuitively, this means we look up “where” (via the key) in $\mathbf{M}$ to find relevant information and then retrieve it (via the value). 
To enable cross attention, the input embeddings $\mathbf{E} \in \mathbb{R}^{T \times d}$ (where $T$ is the sequence length) and memory bank $\mathbf{M} \in \mathbb{R}^{N \times d}$ are projected into query ($\mathbf{Q}$), key ($\mathbf{K}$), and value ($\mathbf{V}$) spaces:
\begin{equation}
\mathbf{Q} = \mathbf{E}_t \mathbf{W}^Q, \quad \mathbf{K} = \mathbf{M}_t \mathbf{W}^K, \quad \mathbf{V} = \mathbf{M}_t \mathbf{W}^V,
\end{equation}
where $\mathbf{W}^Q, \mathbf{W}^K, \mathbf{W}^V \in \mathbb{R}^{d \times d}$ are learnable projection matrices, and $t$ stands for decoder block $t$.

The attention scores are computed as the scaled dot product of the query and key matrices:
\(
\mathbf{A} = \text{softmax}\left(\frac{\mathbf{Q} \mathbf{K}^\top}{\sqrt{d}}\right),
\)
where $\mathbf{A} \in \mathbb{R}^{T \times N}$ represents the alignment between the input sequence and memory slots. The resultant attention output is 
\(
\mathbf{E}_\text{mem} = \mathbf{A} \mathbf{V},
\)
where $\mathbf{E}_\text{mem} \in \mathbb{R}^{T \times d}$ integrates information from the input and memory. To ensure temporal consistency, causal masking is applied, and optionally, top-$k$ attention is used to retain only the most relevant memory interactions.

To regulate the influence of the memory information (\textcolor{gray}{gray} path in Figure~\ref{fig:lm2_wf}) on the existing attention information flow (\textcolor{pink}{pink} path in Figure~\ref{fig:lm2_wf}), an output gate is introduced. 
The output gate dynamically controls the contribution of the memory retrieval based on the cross attention output \(\mathbf{E}_\text{mem}\):
\begin{equation}
g_\text{out} = \sigma\left(\mathbf{E}_\text{mem} \mathbf{W}_\text{out}\right),
\end{equation}
where $\mathbf{W}_\text{out} \in \mathbb{R}^{d \times d}$ is a learnable parameter matrix, and $\sigma$ is the sigmoid activation function. The gated memory output is then computed as:
\begin{equation}
\mathbf{E}_\text{gated} = g_\text{out} \cdot \mathbf{M}_t.
\end{equation}

The gated memory output is integrated into the standard attention flow of the Transformer decoder through a skip connection. Specifically, the output of the self-attention mechanism, $\mathbf{E}_\text{attn}$, is combined with the gated memory output as
\(
\mathbf{E}_\text{next} = \mathbf{E}_\text{attn} + \mathbf{E}_\text{gated}.
\)
This skip connection ensures that the standard attention output and the memory-augmented features jointly contribute to the next decoder layer. By dynamically gating the memory retrieval and integrating it with the attention flow, \name\ effectively balances the use of memory and contextual information, enhancing its ability to model long-term dependencies while preserving the core Transformer operations.

\subsection{Memory updates}

\begin{wrapfigure}{r}{0.5\textwidth}
    \centering
    \includegraphics[width=\linewidth]{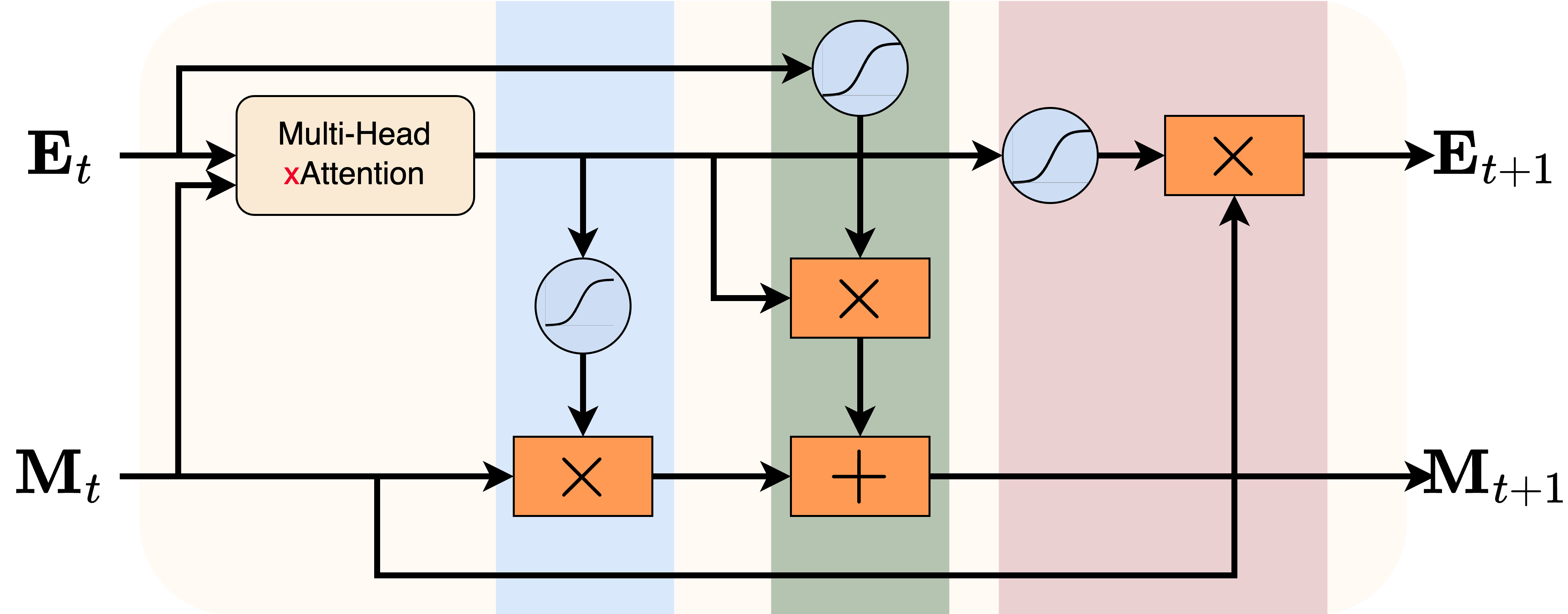}
    \caption{Illustration of how memory module works inside of each decoding block, where \textcolor{blue}{blue}, \textcolor{green}{green}, and \textcolor{rose}{red} box corresponds to forget, input, and output phase.}
    \label{fig:gate}
\end{wrapfigure}

%In this section, we describe how the memory is updated in order to maintain long-term information. 
As illustrated in Figure~\ref{fig:gate}, the update process is divided into three distinct phases: the \emph{input}, \emph{forget}, and \emph{output} (previously described). 
By gating how much new information is introduced and how much old information is discarded, the memory module avoids overwriting crucial long-term facts while also eliminating irrelevant or outdated content when processing long context sequences.

\paragraph{Input Phase}
During the input phase, the model decides how much of the newly computed embeddings (\(\mathbf{E}_\text{mem}\)) to incorporate into the memory. 
To achieve this, first an \emph{input gate} is computed:
\begin{equation}  
g_\text{in} = \sigma\bigl(\mathbf{E}_t \mathbf{W}_\text{in}\bigr),
\end{equation}
where \(\mathbf{W}_\text{in} \in \mathbb{R}^{d \times d}\) is a learnable parameter matrix, 
\(\mathbf{E}_t\) is the current input representation, and \(\sigma\) is the sigmoid activation function. 
This gating mechanism serves as a filter, deciding which relevant information should be “written” into memory, while also preventing the influx of noise or redundant details.

\paragraph{Forgetting Phase}
Once new information is made available during the input phase, the memory must also decide which parts of its existing content to discard. 
This is governed by the \emph{forget gate}:
\begin{equation}
    g_\text{forget} = \sigma\bigl(\mathbf{E}_\text{mem} \mathbf{W}_\text{forget}\bigr),
\end{equation}
where \(\mathbf{W}_\text{forget} \in \mathbb{R}^{d \times d}\). 
By outputting values less than one, the forget gate selectively “erases” memory slots that are no longer relevant, 
allowing the model to focus on more recent or salient information.

\paragraph{Memory Update}
Combining these two gating mechanisms leads to the updated memory state:
\begin{equation}
    \mathbf{M}_{t+1} = g_\text{in} \cdot \tanh(\mathbf{E}_\text{mem}) \;+\; g_\text{forget} \cdot \mathbf{M}_{t},
\end{equation}
where a \(\tanh\) function is applied to keep the new memory content bounded. 
Through these regulated phases, the memory module memorizes the most relevant information and removes outdated details, 
ensuring that it remains both concise and informative over time.

\section{Pre-training \name}
\label{sec:pretraining}

We base our work on the Llama-3 model framework \cite{dubey2024llama}, employing it as the foundation for our Transformer architecture. 
Its architecture comprises 16 decoder blocks, each with a model dimension of 2,048. 
The feed-forward networks within these blocks have an inner dimension of 8,192. The model utilizes 32 attention heads, with 8 dedicated key/value heads. 

Our memory module extends this architecture, consisting of 2,048 memory slots, each with a dimension of 2,048. 
Memory modules are integrated into all 16 decoder blocks, as this configuration empirically achieves the best performance (see Section~\ref{subsec:blocks} for detailed results).
The Llama-3 framework comprises approximately 1.2 billion parameters, with an additional 0.5 billion parameters introduced by the memory module, resulting in a total of 1.7 billion parameters for the \name\ model.

For pre-training, we leverage a high-quality dataset sourced from the SmolLM-Corpus \cite{huggingface_smollm}. 
The dataset is structured into three distinct sections: synthetic test-books and stories, educational web content, and python codes. 
To ensure a focused evaluation on language tasks, we exclude Python sample training data from this process. 
The specific details of the training dataset are outlined as follows:
\textbf{Synthetic Textbooks and Stories}: Generated using advanced language models to cover a wide range of topics, providing 28 billion tokens of diverse educational content.
\textbf{Educational Web Content}: Filtered and deduplicated web pages from FineWeb-Edu \cite{penedo2024fineweb}, contributing 220 billion tokens of high-quality educational material.

% With this memory mechanism, we aim to enhance both the training efficiency and adaptability of the model, enabling it to better handle diverse downstream tasks.
\begin{table*}[t!]
\centering
\caption{Performance on the BABILong dataset: All models are evaluated on various context lengths ranging from 0K, 1K, 2K, and 4K to an aggregated average length of $\geq 8K$. Qa stands for various subsets. 
Due to page limits, we aggregate the results for 8K, 16K, 32K, 64K, and 128K into a single metric, with detailed results provided in Appendix~\ref{sec:babilong_results}.}
\begin{tabular}{cccccccccccc}
\hline
\multicolumn{1}{c|}{model}              & qa1           & qa2           & qa3           & qa4           & qa5           & qa6           & qa7           & qa8           & qa9           & \multicolumn{1}{c|}{qa10}          & Avg.          \\ \hline
\multicolumn{12}{c}{0K}                                                                                                                                                                                                                      \\ \hline
\multicolumn{1}{c|}{Llama-3.2-1.2B}     & 54.0          & 25.0          & 29.0          & 62.0          & 59.0          & 49.0          & 14.0          & 52.0          & 41.0          & \multicolumn{1}{c|}{22.0}          & 40.7          \\
\multicolumn{1}{c|}{vanilla-Llama-1.7B} & 86.0          & 57.0          & 46.0          & 59.0          & 85.0          & 83.0          & 95.0          & 79.0          & 83.0          & \multicolumn{1}{c|}{77.0}          & 75.0          \\
\multicolumn{1}{c|}{RMT-1.7B}           & 85.0          & 49.0          & 49.0          & 81.0          & 95.0          & 84.0          & 82.0          & 78.0          & 85.0          & \multicolumn{1}{c|}{76.0}          & 76.4          \\
\multicolumn{1}{c|}{LM2-1.7B}           & \textbf{99.0} & \textbf{89.0} & \textbf{70.0} & \textbf{88.0} & \textbf{98.0} & \textbf{95.0} & \textbf{96.0} & \textbf{97.0} & \textbf{99.0} & \multicolumn{1}{c|}{\textbf{94.0}} & \textbf{92.5} \\ \hline
\multicolumn{12}{c}{1K}                                                                                                                                                                                                                      \\ \hline
\multicolumn{1}{c|}{Llama-3.2-1.2B}     & 48.0          & 22.0          & 24.0          & 55.0          & 69.0          & 49.0          & 9.0           & 31.0          & 55.0          & \multicolumn{1}{c|}{33.0}          & 39.5          \\
\multicolumn{1}{c|}{Llama-3.2-1.2B-RAG} & 51.0            & 14.0            & 19.0            & 59.0            & 80.0            & 49.0            & 10.0            & 38.0            & 40.0            & \multicolumn{1}{c|}{46.0}            & 40.6          \\
\multicolumn{1}{c|}{vanilla-Llama-1.7B} & 31.0          & 21.0          & 44.0          & 43.0          & 71.0          & 60.0          & 71.0          & 40.0          & 67.0          & \multicolumn{1}{c|}{58.0}          & 50.6          \\
\multicolumn{1}{c|}{RMT-1.7B}           & 35.0          & 26.0          & 29.0          & 33.0          & 61.0          & 50.0          & 83.0          & 41.0          & 68.0          & \multicolumn{1}{c|}{53.0}          & 47.9          \\
\multicolumn{1}{c|}{LM2-1.7B}           & \textbf{85.0} & \textbf{59.0} & \textbf{72.0} & \textbf{68.0} & \textbf{91.0} & \textbf{84.0} & \textbf{96.0} & \textbf{69.0} & \textbf{82.0} & \multicolumn{1}{c|}{\textbf{77.0}} & \textbf{78.3} \\ \hline
\multicolumn{12}{c}{2K}                                                                                                                                                                                                                      \\ \hline
\multicolumn{1}{c|}{Llama-3.2-1.2B}     & 44.0          & 18.0          & 19.0          & \textbf{50.0} & 64.0          & 52.0          & 18.0          & 24.0          & 55.0          & \multicolumn{1}{c|}{42.0}          & 38.6          \\
\multicolumn{1}{c|}{Llama-3.2-1.2B-RAG} & 52.0            & 11.0          & 12.0          & 49.0          & 75.0          & 48.0          & 5.0           & 33.0          & 50.0          & \multicolumn{1}{c|}{43.0}          & 37.8          \\
\multicolumn{1}{c|}{vanilla-Llama-1.7B} & 25.0          & 22.0          & 37.0          & 34.0          & 58.0          & 60.0          & 65.0          & 38.0          & 66.0          & \multicolumn{1}{c|}{58.0}          & 46.3          \\
\multicolumn{1}{c|}{RMT-1.7B}           & 44.0          & 21.0          & 43.0          & 41.0          & 79.0          & 47.0          & 78.0          & 41.0          & 69.0          & \multicolumn{1}{c|}{51.0}          & 51.4          \\
\multicolumn{1}{c|}{LM2-1.7B}           & \textbf{58.0} & \textbf{43.0} & \textbf{64.0} & 43.0          & \textbf{87.0} & \textbf{73.0} & \textbf{93.0} & \textbf{53.0} & \textbf{75.0} & \multicolumn{1}{c|}{\textbf{69.0}} & \textbf{65.8} \\ \hline
\multicolumn{12}{c}{4K}                                                                                                                                                                                                                      \\ \hline
\multicolumn{1}{c|}{Llama-3.2-1.2B}     & 37.0          & 16.0          & 25.0          & 56.0          & 56.0          & 50.0          & 14.0          & 27.0          & 55.0          & \multicolumn{1}{c|}{32.0}          & 36.8          \\
\multicolumn{1}{c|}{Llama-3.2-1.2B-RAG} & 47.0            & 3.0             & 16.0            & \textbf{58.0}   & 68.0            & 58.0            & 3.0             & 36.0            & 45.0            & \multicolumn{1}{c|}{39.0}            & 37.3          \\
\multicolumn{1}{c|}{vanilla-Llama-1.7B} & 21.0          & 18.0          & 38.0          & 28.0          & 55.0          & 61.0          & 64.0          & 35.0          & 49.0          & \multicolumn{1}{c|}{53.0}          & 42.2          \\
\multicolumn{1}{c|}{RMT-1.7B}           & 24.0          & 20.0          & 22.0          & 24.0          & 28.0          & 46.0          & 75.0          & 35.0          & \textbf{65.0} & \multicolumn{1}{c|}{45.0}          & 38.4          \\
\multicolumn{1}{c|}{LM2-1.7B}           & \textbf{46.0} & \textbf{37.0} & \textbf{48.0} & 34.0          & \textbf{78.0} & \textbf{66.0} & \textbf{93.0} & \textbf{45.0} & 62.0          & \multicolumn{1}{c|}{\textbf{50.0}} & \textbf{55.9} \\ \hline
\multicolumn{12}{c}{AVG. Length $\geq$8K}                                                                                                                                                                                                    \\ \hline
\multicolumn{1}{c|}{Llama-3.2-1.2B}     & 19.0          & 8.0           & 17.8          & 27.3          & 36.5          & 49            & 21.3          & 12.8          & 48.0          & \multicolumn{1}{c|}{41.8}          & 28.2          \\
\multicolumn{1}{c|}{Llama-3.2-1.2B-RAG} & 29.3          & 1.0             & 5.0             & \textbf{55.8} & \textbf{72.0}            & \textbf{49.8} & 4.8           & 22.8          & 46.3          & \multicolumn{1}{c|}{36.8}          & 32.3          \\
\multicolumn{1}{c|}{vanilla-Llama-1.7B} & 11.3          & 15.0          & 21.3          & 14.5          & 31.0            & 44.0            & 63.0            & 33.5          & 42.0            & \multicolumn{1}{c|}{36.3}          & 31.2          \\
\multicolumn{1}{c|}{RMT-1.7B}           & 17.5          & 14.5          & 20.5          & 22.5          & 20.3          & 47.0          & 73.3          & 34.5          & 62.5          & \multicolumn{1}{c|}{\textbf{43.0}}   & 35.5          \\
\multicolumn{1}{c|}{LM2-1.7B}           & \textbf{23.8} & \textbf{15.0} & \textbf{24.5} & 24.0            & 38.8 & 47.3          & \textbf{92.8} & \textbf{37.0}   & \textbf{53.8} & \multicolumn{1}{c|}{42.0}            & \textbf{39.9} \\ \hline
\end{tabular}
\label{tab:babilong}
\end{table*}

\section{Experiments}

We design our experiments to answer the following questions:
\textbf{Q1:} How does \name\ perform in memory tasks?
\textbf{Q2:} Does \name\ harm the performance in general tasks?
\textbf{Q3:} Do we need to include the memory module in all decoder blocks?
\textbf{Q4:} What is stored in the memory bank?
\textbf{Q5:} How is the memory module updated at test-time?

To evaluate \name, we compare its performance against the following baselines:  
\textbf{vanilla-Llama-1.7B:} The Llama 3.2 architecture, pre-trained from scratch on the same datasets as \name. We scale this model to 1.7 billion parameters for a fair comparison.  
\textbf{RMT-1.7B:} Recurrent Memory Transformer (RMT)~\cite{bulatov2022recurrentmemorytransformer} is a memory-augmented framework that generates memory tokens, serving as an additional module built on top of existing LLMs. 
We use the LLaMA-1.7B model as the backbone and fine-tune it on the bAbI training dataset \cite{westonBCM15}, following the methodology outlined in \citet{kuratov2024babilong} and \citet{ko2024memreasoner}.
% \textbf{MemReasoner:} A model proposed by~\cite{ko2024memreasoner}, which employs memory ordering techniques for retrieving factors in long context reasoning tasks. 
\textbf{Llama-3.2-1.2B:} To show case the effectiveness of \name\, we also compared the model against the original model trained by Meta, with the same total number of pure Transformer parameters (1.2B), but trained on far more high-quality tokens.
\textbf{Llama-3.2-1.2B-RAG:} Lastly, we compare with a version of Llama with retrieval-augmented generation (RAG) to better handle long context problems.

\subsection{Performance on Memory Tasks}

\paragraph{BABILong} The BABILong dataset \cite{kuratov2024babilong} extends bAbI benchmark \cite{westonBCM15} by incorporating significantly longer contexts and more intricate queries, thus demanding advanced memory capabilities and multi-step reasoning. 
By increasing both contextual and computational challenges, BABILong offers a rigorous evaluation benchmark for testing memory-augmented models. 
% Together, these datasets play a pivotal role in advancing research on reasoning, generalization, and contextual understanding in natural language processing.

Table~\ref{tab:babilong} presents a comparison of our model against the baselines on the BABILong dataset.
We report results across multiple context lengths, from 0K context-length, which is identical to bAbI dataset, to the maximum context length of 128K, which is the target context-length of the backbone Llama-3.2 model.
From this table, we observe several key findings as follows:

\begin{wrapfigure}{r}{0.5\textwidth}
    \centering
    \includegraphics[width=\linewidth]{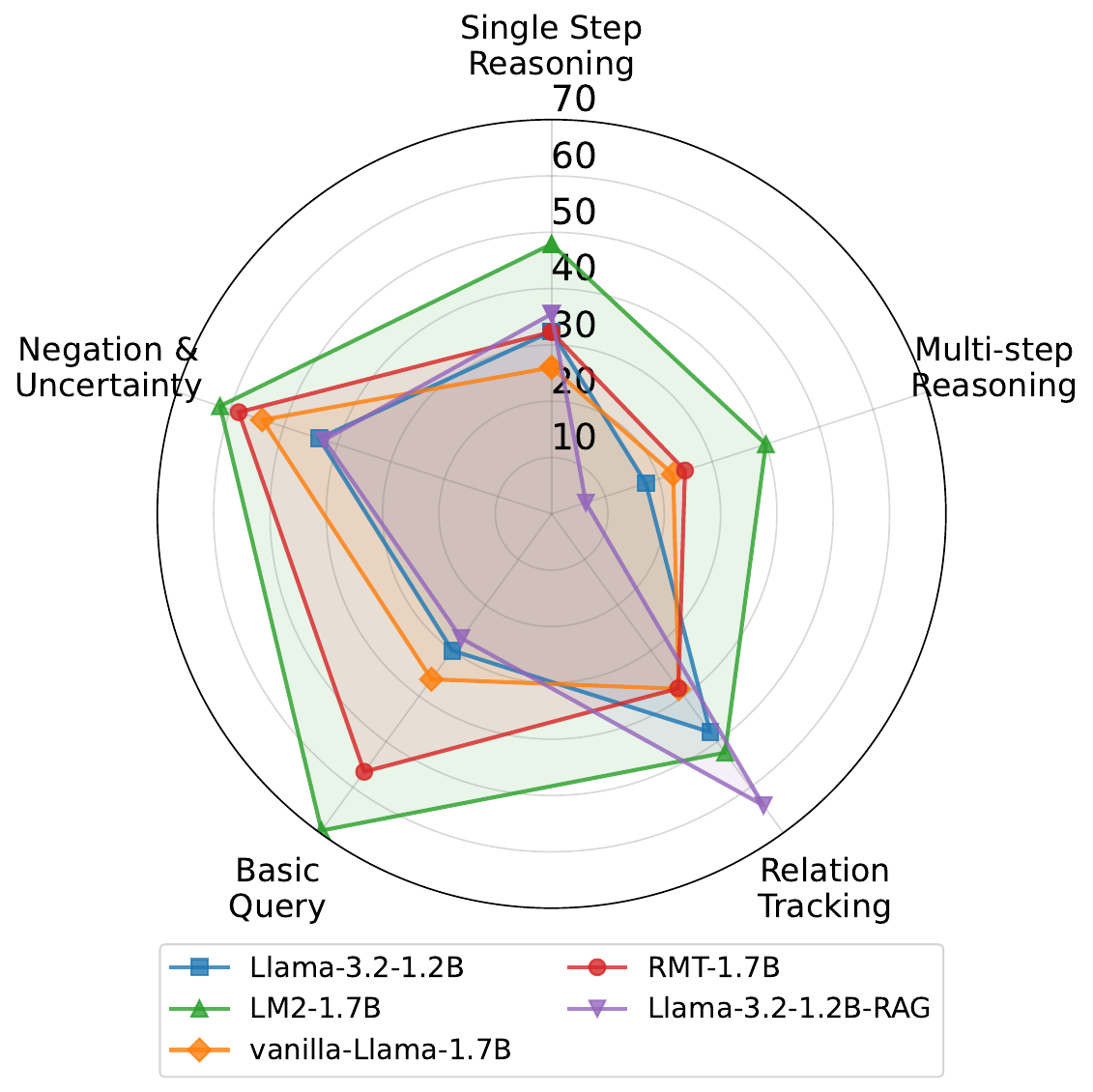}
    \caption{Performance on BABILong benchmark with different capabilities.}
    \label{fig:babilong_radar}
\end{wrapfigure}

\paragraph{Performance at bAbI benchmark (0K).}
Without additional context, LM2-1.7B achieves the highest average accuracy of 92.5\%, surpassing Llama-3.2-1.2B, vanilla-LLama-1.7B, and RMT-1.7B, which average results are 40.7\%, 75.0\% and 76.4\%, respectively.
Because Llama-3.2-1.2B-RAG is designed for retrieval-augmented generation and evaluated only at longer contexts, it is not included in the 0K setting. 
This suggests that LM2’s underlying modeling improvements enhance its core reasoning ability.

\paragraph{Performance at Long Context Lengths (1K--4K).}
As context length increases, performance generally degrades for all models, but LM2-1.7B maintains a noticeable improvement over both standard and retrieval-augmented Llama variants and RMT.
For instance, at 4K, LM2-1.7B’s average accuracy (55.9\%) is higher than Llama-3.2-1.2B, vanilla-LLama-1.7B, and RMT-1.7B, which average results are 36.8\%, 42.2\% and 48.4\%, respectively.
This gap underscores LM2’s effectiveness for long-term memory ranging from 1K to 4K.

\paragraph{Performance at Long Contexts (8K--128K).}
Although all models exhibit some accuracy decline at these extreme long context lengths, LM2-1.7B remains robust. 
% Notably, LM2-1.7B achieves 92.8\% on QA7 and 53.8\% on QA9, significantly outperforming the other models. 
RMT-1.7B shows reasonable robustness, yet still falls short of LM2-1.7B on most tasks. 
RAG methods demonstrate some improvements over the baseline Llama, but still falls behind memory-based methods.
These results highlight LM2’s ability to handle long context problems where Transformer-based models struggle.

\paragraph{Performance at Different Reasoning Types} To further understand how \name\ performs in different reasoning scenarios, we group the BABILong dataset into five categories: (1) Single-step Reasoning (qa1), (2) Multi-step Reasoning (qa2–3), (3) Relation Tracking (qa4–5), (iv) Basic Queries (qa6–8), and (v) Negation \& Uncertainty (qa9–10). 
Figure~\ref{fig:babilong_radar} depicts the results in a radar chart, where higher values indicate better performance.
Across nearly all task categories except for \emph{Relation Tracking}, LM2-1.7B demonstrates the best performance. 
Notably, \name\ outperforms the other methods on both single and multi-step reasoning, indicating that it can handle more complex, multi-hop inferences and direct fact retrieval with fewer errors. 
The improvement margin is larger for Basic Queries, Single-Step Reasoning, and Multi-step Reasoning, suggesting that LM2 has strong abilities to retrieve long-term facts and apply them in complex reasoning tasks.
The marginally lower performance on Relation Tracking can be attributed to RAG’s approach of chunking the context into smaller, more focused “documents” and retrieving only the most relevant pieces at inference time. 
RAG makes it much easier to precisely identify which facts are associated with the queried relationship, thus serving as an extremely strong baseline for this task category.

\subsection{Performance on General Benchmarks}

To further evaluate if introducing an extra memory module affects LLMs' general performance, we evaluate the proposed memory-based model, LM2, on the MMLU benchmark \cite{hendryckstest2021}, which tests a broad spectrum of subject areas—STEM, Humanities, Social Sciences, and Others—as well as varied difficulty levels—High School, College, Professional, and General Knowledge. 
Table \ref{tab:mmlu} presents the results of LM2 in comparison to vanilla-Llama and RMT.

% \begin{table}[ht]
\begin{wraptable}{r}{0.5\textwidth}
\centering
\caption{Performance on MMLU dataset. For better visualization, the dataset is categorized on two criteria - subject and difficulty. }
\begin{tabular}{cc|ccc}
\cline{3-5}
                                                                                                 &                                                              & \begin{tabular}[c]{@{}c@{}}vanilla\\ Llama\end{tabular} & RMT  & LM2           \\ \hline
\multicolumn{1}{c|}{\multirow{4}{*}{\begin{tabular}[c]{@{}c@{}}Subject\\ Category\end{tabular}}} & STEM                                                         & 27.2                                                    & 25.7 & \textbf{28.1} \\
\multicolumn{1}{c|}{}                                                                            & Humanities                                                   & 28.7                                                    & 26.7 & \textbf{32.2} \\
\multicolumn{1}{c|}{}                                                                            & Social Sciences                                              & 29.2                                                    & 27.0 & \textbf{31.6} \\
\multicolumn{1}{c|}{}                                                                            & Others                                                       & 27.7                                                    & 27.1 & \textbf{28.0} \\ \hline
\multicolumn{1}{c|}{\multirow{4}{*}{\begin{tabular}[c]{@{}c@{}}Difficulty\\ Level\end{tabular}}} & High School                                                  & 28.8                                                    & 26.5 & \textbf{30.4} \\
\multicolumn{1}{c|}{}                                                                            & College                                                      & 27.7                                                    & 27.1 & \textbf{29.0} \\
\multicolumn{1}{c|}{}                                                                            & Professional                                                 & 27.5                                                    & 26.6 & \textbf{27.6} \\
\multicolumn{1}{c|}{}                                                                            & \begin{tabular}[c]{@{}c@{}}General \\ Knowledge\end{tabular} & 27.2                                                    & 25.6 & \textbf{28.5} \\ \hline
\multicolumn{2}{c|}{Average}                                                                                                                                    & 28.0                                                      & 26.5 & \textbf{29.4} \\ \hline
\end{tabular}
\label{tab:mmlu}
\end{wraptable}

Overall, LM2 demonstrates a clear performance gain, improving the average accuracy of vanilla-Llama from 28.0\% to 29.4\%. 
On the contrary, despite sharing the same pre-trained model, RMT degrades the performance of vanilla-Llama to 26.5\%.
Notably, \name\ achieves substantial gains in Humanities and Social Sciences, where LM2 surpasses vanilla-Llama by 3.5\% and 2.4\%, respectively. 
These categories often involve context-rich questions, suggesting that LM2’s memory-based approach is advantageous for retaining and leveraging more nuanced and interconnected information. 
Meanwhile, LM2 also sustains competitive performance in STEM and Others, indicating its robustness beyond highly specialized domains.

These results illustrate that LM2 overcomes the drawback associated with memory-augmented models: performance degradation on more general tasks. 
Current memory-based architectures are carefully designed for memory tasks, weakening their ability to general LLM tasks. 
However, LM2’s performance on all categories of MMLU dataset indicates that the proposed memory mechanism does not impede its general applicability.

\subsection{Impact of memory modules}
\label{subsec:blocks}

\begin{figure}[t]
    \centering
    \includegraphics[width=0.7\linewidth]{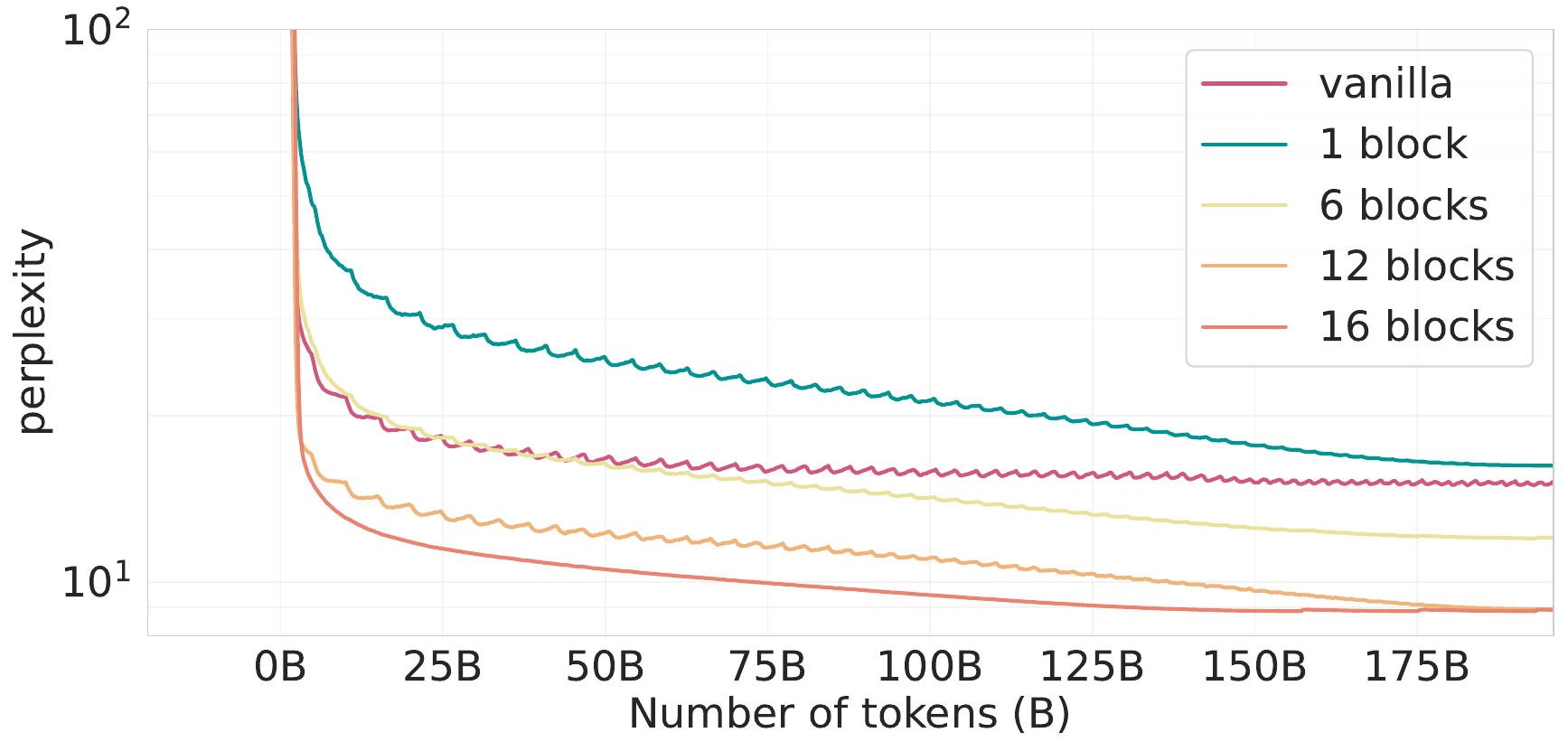}
    \caption{We evaluate variations of integrating memory within the decoder blocks. The number indicates how many of the initial decoder blocks include the memory module, as we found that the order of implementing memory modules does not affect performance.}
    \label{fig:abs_blocks}
\end{figure}

We evaluate the effectiveness of proposed memory modules using perplexity as the primary metric across varying numbers of training tokens (measured in billions).
Figure~\ref{fig:abs_blocks} illustrates the perplexity trends for the baseline vanilla-Llama and \name\ with varying degrees of memory integration (i.e., 1, 6, 12, and 16 blocks), where 16 is the maximum number of blocks used in Llama-3.2-1B.

The results demonstrate that integrating memory information more extensively throughout the decoder leads to improved model performance. 
Specifically, implementing the memory module in only the first block achieves similar results to the vanilla Llama, but with slower convergence. This suggests that introducing a single memory flow does not degrade overall performance but may slow down training because of extra memory optimization.
\begin{wrapfigure}[25]{r}{0.5\textwidth}
    \centering
    \includegraphics[width=\linewidth]{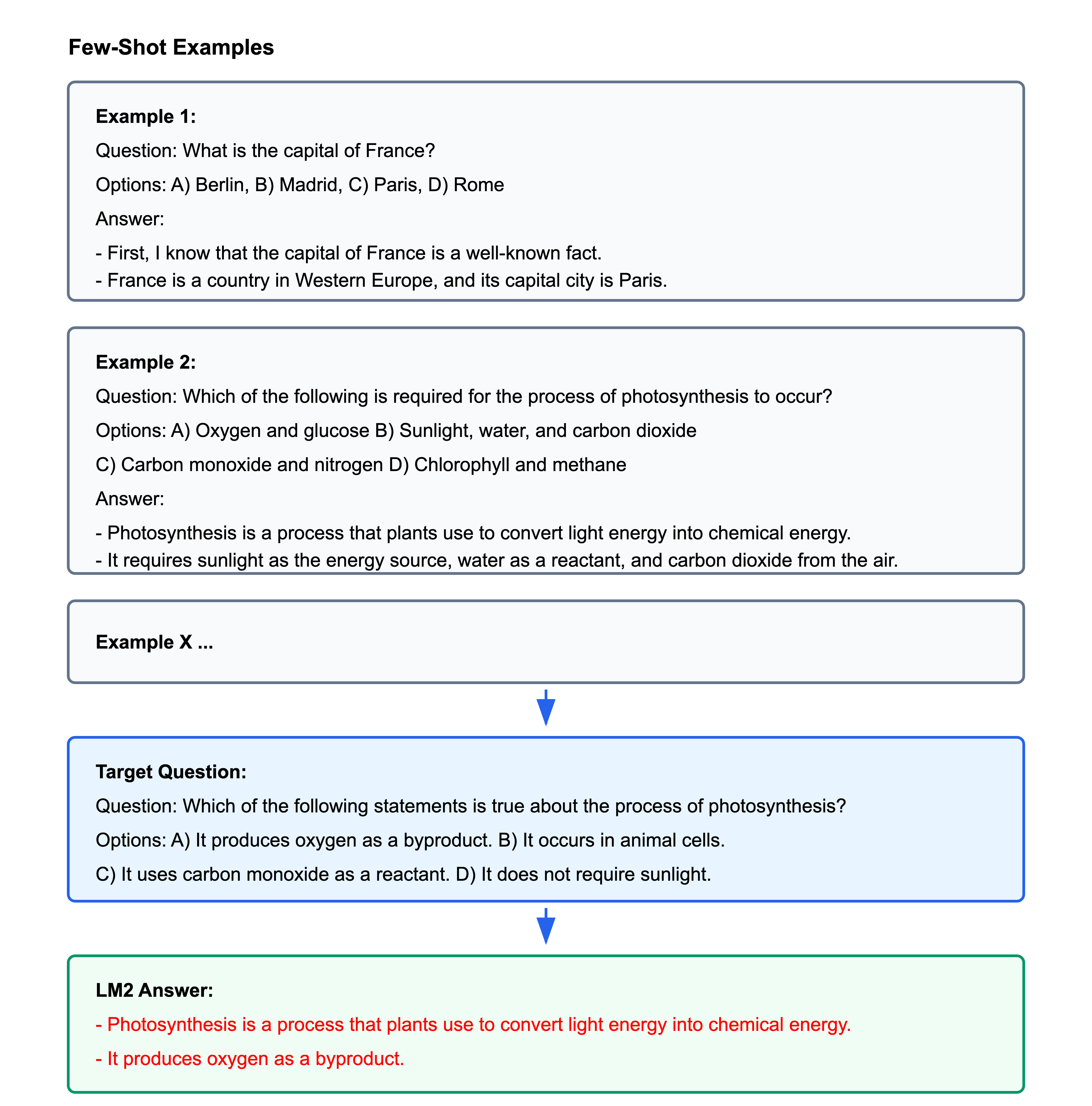}
    \caption{We sample a question from MMLU to test the \name\ in a few-shot fashion. To study how the memory module focuses on relevant information, we place useful information inside one of the few-shot examples.}
    \label{fig:few-shot-example}
\end{wrapfigure}
In contrast, incorporating more memory flows, such as in the 6-block configuration, leads to lower perplexity, highlighting the effectiveness of the proposed memory flow design.
The 16-block configuration significantly outperforms the limited 1-block integration, validating that the proposed memory flow is highly advantageous for reducing perplexity and enhancing the overall capabilities of the model.

% \begin{wrapfigure}[28]{r}{0.5\textwidth}
%     \centering
%     \includegraphics[width=0.9\linewidth]{figs/perplexity_plot.pdf}
%     \caption{We evaluate variations of integrating memory within the decoder blocks. The number indicates how many of the initial decoder blocks include the memory module, as we found that the order of implementing memory modules does not affect performance.}
%     \label{fig:abs_blocks}
% \end{wrapfigure}

\subsection{Analysis of Memory Representations}

To gain deeper insights into the information encoded within the memory module, we utilize the Neuron Explainer method \cite{bills2023language}. 
It generates natural language explanations of neuron behavior, simulates activations using these descriptions, and evaluates their accuracy through predictive scoring. 
We utilize this approach to explain the latent representations of specific memory slots, which helps understand how these slots process and retain task-relevant information. 
By analyzing activations within the memory module, the Neuron Explainer identifies patterns in latent representations of each memory slot, mapping them to specific elements of the input text. 

We evaluate \name\ using the input text illustrated in Figure~\ref{fig:few-shot-example}. 
Subsequently, we identify and rank the most relevant memory slots, selecting two for sampling (slots 1679 and 1684) along with one of the least relevant memory slot (slot 1). 
Utilizing the neuron explainer, we investigate the relevance rationales.

\begin{explain}{Memory Slot 1679}{m1}
    This memory slot's representations for this specific input text suggest that the memory module's focus is likely on detecting factual information, question and answer structures.
\end{explain}

These observations suggest that \textbf{Memory Slot 1679} specializes in retrieving and synthesizing factual information for the target question, functioning as a repository for domain-specific knowledge and structured reasoning.

\begin{explain}{Memory Slot 1684}{m2}
    This representations in this memory slot is designed to focus on specific elements within the input text, as evidenced by the pattern in the memory bank.    
\end{explain}

\textbf{Memory Slot 1684}, in contrast, demonstrated a focus on structural elements within the input text. Its activations aligned closely with linguistic markers and contextual cues, such as “Options:” or “Answer:”. 
This behavior implies that Memory Slot 1684 facilitates the model's comprehension of input organization, enabling effective parsing of complex instruction formats and multi-part structures. 

\begin{explain}{Memory Slot 1}{m3}
    The representations in this memory slot for the provided input text are primarily negative. This suggests that the module is not detecting the specific aspects it was designed to recognize in the input text.   
\end{explain}

\textbf{Memory Slot 1}, on the other hand, showed predominantly negative activations across the input text, indicating minimal engagement with the task-specific content. 

These findings underline the importance of memory modules in gathering information for the generation tasks.

\subsection{Test-time memory adaptations}

\begin{figure}[ht]
    \centering
    \begin{subfigure}[b]{0.45\linewidth}
        \centering
        \includegraphics[width=\linewidth]{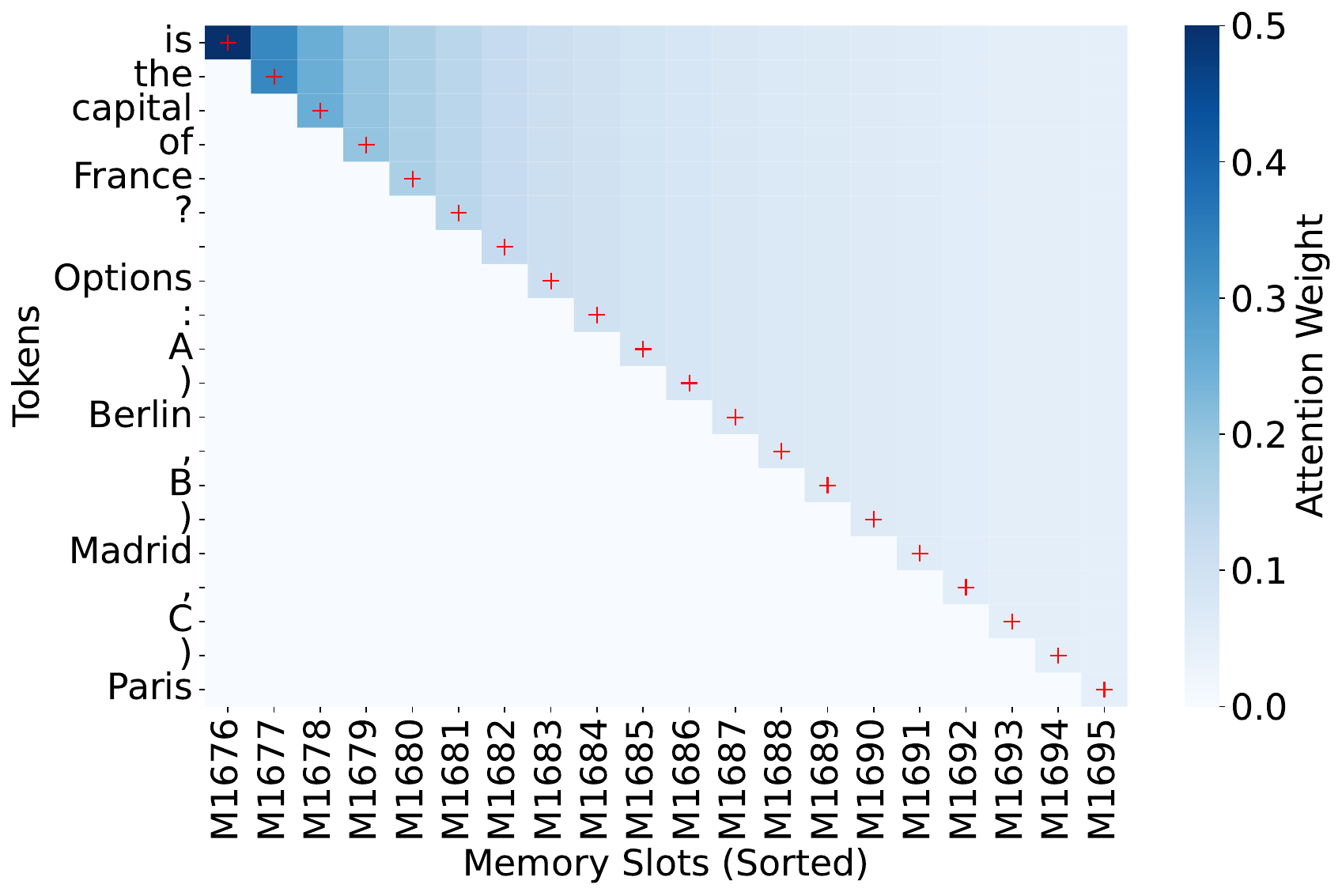}
        \caption{Cross-attention heatmaps before memory update.}
        \label{fig:xattn-before}
    \end{subfigure}
    \hfill
    \begin{subfigure}[b]{0.45\linewidth}
        \centering
        \includegraphics[width=\linewidth]{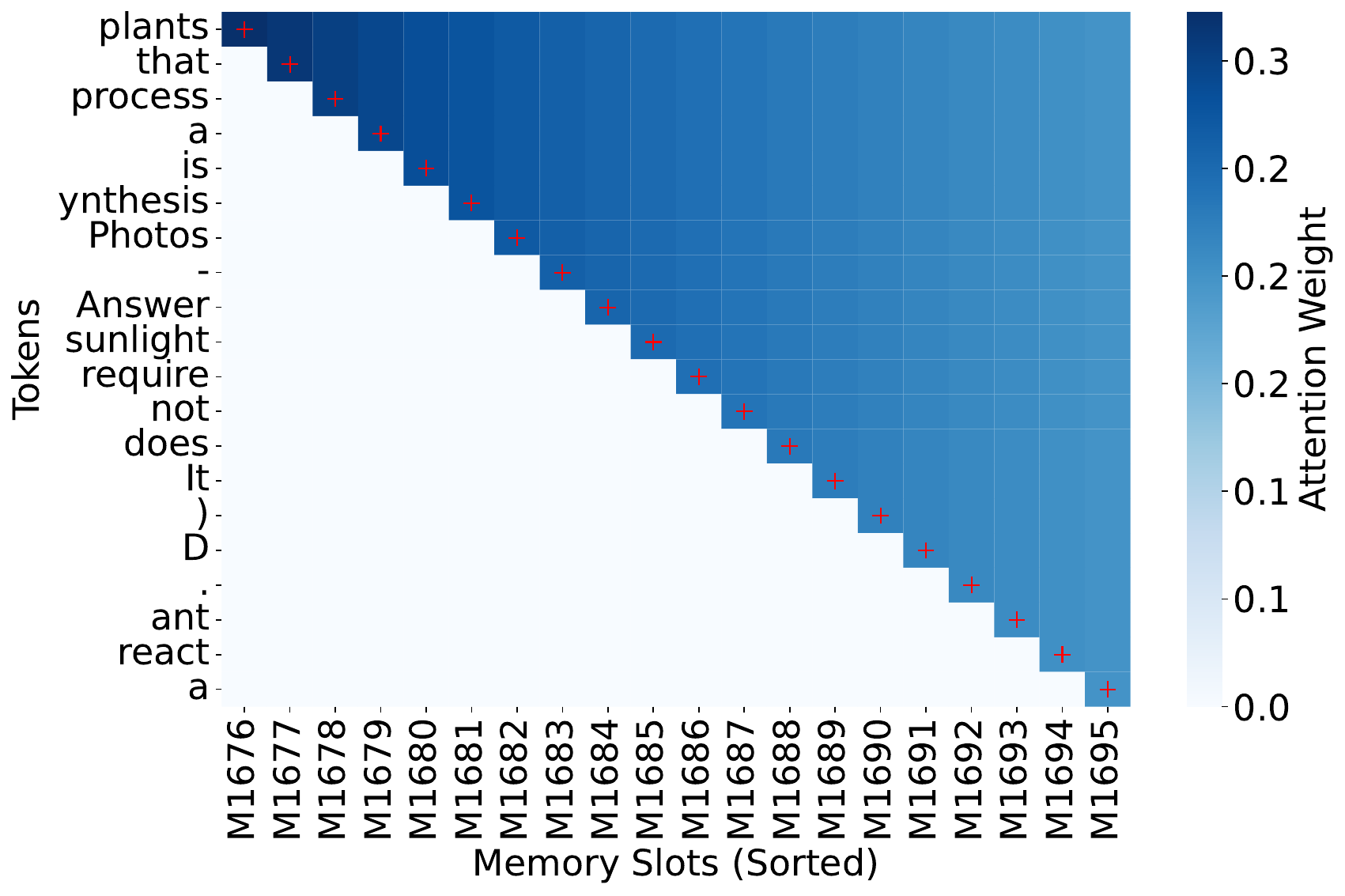}
        \caption{Cross-attention heatmaps after memory update.}
        \label{fig:xattn-after}
    \end{subfigure}
    \caption{Cross-attention heatmaps between input tokens and memory. The x-axis shows the memory slots sorted by slot number. The y-axis shows the most attended tokens. Diagonal attentions are marked with ``+''.}
    \label{fig:xattn}
\end{figure}

We further investigate how memory updates influence model generation during test time. 
To explore this, we analyze the example illustrated in Figure~\ref{fig:few-shot-example}. 
Cross attention heatmaps, presented in Figure~\ref{fig:xattn}, provide key insights into these memory updates.

Figure~\ref{fig:xattn-before} shows the cross attention heatmap prior to memory updates. 
In this figure, tokens such as “France” and “Paris” strongly engage with the memory. These tokens do not pertain specifically to the target question about photosynthesis. Instead, on the first pass, memory initially focuses on the structure of question as well as identifying factual information.

Next, we examine the memory heatmap after various inference update steps (one inference step corresponds to a single forward pass for one token). 
As depicted in Figure~\ref{fig:xattn-after}, the tokens attended to by the memory slots shift toward those relevant to the target question.
Since cross attention exclusively computes the relationships between input tokens and memory, this shift reflects the influence of test-time memory updates. 
These changes highlight the adaptive nature of memory during inference.

\section{Related Work}
\paragraph{Memory augmented Transformers}
Various methods have been proposed to augment Transformers with memory. One direction is to optimize the attention mechanisms and use some global representations acting as memory points to ensure input coverage. Models like Longformer \cite{DBLP:journals/corr/abs-2004-05150}, Big Bird \cite{DBLP:journals/corr/abs-2007-14062}, GMAT \cite{DBLP:journals/corr/abs-2006-03274} and Extended Transformer Construction \cite{DBLP:journals/corr/abs-2004-08483} all proposed some sparse attention mechanisms to reduce the quadratic dependency of self-attention to linear and introduced global tokens to encode the information from the entire sequence.

Another line of work introduces memorization capabilities to Transformers through recurrence. Transformer-XL \cite{DBLP:journals/corr/abs-1901-02860} addresses the limitation of fixed-length context by introducing segment-level recurrence and relative position encodings. 
% It processes input sequences segment by segment, caching hidden states from previous parts to extend the context window efficiently without recomputing the entire sequence, improving performance on long-sequence tasks. 
However, during training, gradients are restricted to individual segments, limiting the model's ability to capture long-term temporal dependencies. 
% Moreover, using all hidden states from previous segments as memory for each layer significantly increases computational and memory demands during the forward pass. 
Recurrent Memory Transformer (RMT)~\cite{bulatov2022recurrentmemorytransformer} mitigates these limitations by introducing a more efficient memory mechanism. It adds recurrence to Transformers via a small number of special overlapping memory tokens between segments of long sequences, enabling gradients to propagate across them while significantly reducing memory usage. 
RMT outperforms Transformer-XL for sequence processing tasks and is on par with Transformer-XL on language modeling, but requires less memory.
Associative RMT (ARMT) \cite{rodkin2024associativerecurrentmemorytransformer} is a follow-up to RMT that addresses its time complexity issues. 
Similarly, MemReasoner \cite{ko2024memreasoner} introduces a memory-augmented LLM architecture designed for temporal reasoning. However, as demonstrated by \citet{kuratov2024babilong} and \citet{ko2024memreasoner}, RMT continues to outperform these subsequent models, maintaining its status as the state-of-the-art (SOTA) method. 
Therefore, we primarily consider RMT as the SOTA memory-based model and compare \name\ against it.

\paragraph{Retrieval-Augmented Generation (RAG)}
Retrieval-Augmented Generation (RAG) \cite{DBLP:journals/corr/abs-2005-11401} is a popular solution for language models to handle large amounts of text. The core architecture of RAG comprises a retriever module that identifies relevant information from a knowledge base, ensuring that the input to the generative model remains within the token limit while filtering out irrelevant noise, thereby improving efficiency and response quality.
% Hybrid search is a common approach for retrievers, combining dense embeddings generated by bi-encoders for semantic matching with sparse methods like BM25 for precise lexical retrieval, balancing contextual understanding and term accuracy. Following the retrieval phase, a re-ranking module, often implemented using a cross-encoder, can be introduced to filter out top-K from the retrieved documents, forming a two-stage retrieval pipeline.
While Retrieval-Augmented Generation (RAG) has proven effective for many tasks, it struggles with some complicated tasks like multi-hop question-answering \cite{mavi2024multihopquestionanswering}, which require retrieving and reasoning over multiple interconnected pieces of evidence.

\section{Conclusion}

In this paper, we introduced \fullname\ (\name), a memory-augmented Transformer architecture designed to address long context reasoning challenges. 
The key innovation is the memory module, integrated inside the decoder blocks, which augments the model with additional memory information while also updating itself. 
Empirical results on the BABILong benchmark highlights \name’s advantages on various long context tasks.
On average across tasks, \name\ outperforms the SOTA memory-augmented RMT model by 37.1\%, and a non-memory baseline Llama-3.2 model by 86.3\%. 
Furthermore, \name\ achieves improvement over baselines on the MMLU benchmark, evidencing that its memory module does not degrade performance on general tasks. 
% Analyses of the memory activations and test-time updates further confirm that \name\ identifies and retains context-relevant information, adaptively adjusting its stored representations to changing inputs. 
Overall, these findings underscore the importance of explicit memory mechanisms, and lay a foundation for further research on integrating long-term memory into large language models.

\clearpage
\newpage
\bibliographystyle{plainnat}
\bibliography{references}

\begin{thebibliography}{26}
\providecommand{\natexlab}[1]{#1}
\providecommand{\url}[1]{\texttt{#1}}
\expandafter\ifx\csname urlstyle\endcsname\relax
  \providecommand{\doi}[1]{doi: #1}\else
  \providecommand{\doi}{doi: \begingroup \urlstyle{rm}\Url}\fi

\bibitem[Ainslie et~al.(2020)Ainslie, Onta{\~{n}}{\'{o}}n, Alberti, Pham, Ravula, and Sanghai]{DBLP:journals/corr/abs-2004-08483}
Joshua Ainslie, Santiago Onta{\~{n}}{\'{o}}n, Chris Alberti, Philip Pham, Anirudh Ravula, and Sumit Sanghai.
\newblock {ETC:} encoding long and structured data in transformers.
\newblock \emph{CoRR}, abs/2004.08483, 2020.
\newblock URL \url{https://arxiv.org/abs/2004.08483}.

\bibitem[Beltagy et~al.(2020)Beltagy, Peters, and Cohan]{DBLP:journals/corr/abs-2004-05150}
Iz~Beltagy, Matthew~E. Peters, and Arman Cohan.
\newblock Longformer: The long-document transformer.
\newblock \emph{CoRR}, abs/2004.05150, 2020.
\newblock URL \url{https://arxiv.org/abs/2004.05150}.

\bibitem[Bills et~al.(2023)Bills, Cammarata, Mossing, Tillman, Gao, Goh, Sutskever, Leike, Wu, and Saunders]{bills2023language}
Steven Bills, Nick Cammarata, Dan Mossing, Henk Tillman, Leo Gao, Gabriel Goh, Ilya Sutskever, Jan Leike, Jeff Wu, and William Saunders.
\newblock Language models can explain neurons in language models.
\newblock \url{https://openaipublic.blob.core.windows.net/neuron-explainer/paper/index.html}, 2023.

\bibitem[Brown et~al.(2020)Brown, Mann, Ryder, Subbiah, Kaplan, Dhariwal, Neelakantan, Shyam, Sastry, Askell, et~al.]{brown2020language}
Tom Brown, Benjamin Mann, Nick Ryder, Melanie Subbiah, Jared~D Kaplan, Prafulla Dhariwal, Arvind Neelakantan, Pranav Shyam, Girish Sastry, Amanda Askell, et~al.
\newblock Language models are few-shot learners.
\newblock \emph{Advances in neural information processing systems}, 33:\penalty0 1877--1901, 2020.

\bibitem[Bulatov et~al.(2022)Bulatov, Kuratov, and Burtsev]{bulatov2022recurrentmemorytransformer}
Aydar Bulatov, Yuri Kuratov, and Mikhail~S. Burtsev.
\newblock Recurrent memory transformer, 2022.
\newblock URL \url{https://arxiv.org/abs/2207.06881}.

\bibitem[Dai et~al.(2019)Dai, Yang, Yang, Carbonell, Le, and Salakhutdinov]{DBLP:journals/corr/abs-1901-02860}
Zihang Dai, Zhilin Yang, Yiming Yang, Jaime~G. Carbonell, Quoc~V. Le, and Ruslan Salakhutdinov.
\newblock Transformer-xl: Attentive language models beyond a fixed-length context.
\newblock \emph{CoRR}, abs/1901.02860, 2019.
\newblock URL \url{http://arxiv.org/abs/1901.02860}.

\bibitem[Dooley(2007)]{DooleyArchivalAdvantage}
Jackie Dooley.
\newblock The archival advantage: Integrating archival expertise into management of born-digital library materials.
\newblock \emph{Archival Science Special Issue on Archiving Research Data}, 7\penalty0 (1), March 2007.

\bibitem[Dosovitskiy(2020)]{dosovitskiy2020image}
Alexey Dosovitskiy.
\newblock An image is worth 16x16 words: Transformers for image recognition at scale.
\newblock \emph{arXiv preprint arXiv:2010.11929}, 2020.

\bibitem[Dubey et~al.(2024)Dubey, Jauhri, Pandey, Kadian, Al-Dahle, Letman, Mathur, Schelten, Yang, Fan, et~al.]{dubey2024llama}
Abhimanyu Dubey, Abhinav Jauhri, Abhinav Pandey, Abhishek Kadian, Ahmad Al-Dahle, Aiesha Letman, Akhil Mathur, Alan Schelten, Amy Yang, Angela Fan, et~al.
\newblock The llama 3 herd of models.
\newblock \emph{arXiv preprint arXiv:2407.21783}, 2024.

\bibitem[Gupta and Berant(2020)]{DBLP:journals/corr/abs-2006-03274}
Ankit Gupta and Jonathan Berant.
\newblock {GMAT:} global memory augmentation for transformers.
\newblock \emph{CoRR}, abs/2006.03274, 2020.
\newblock URL \url{https://arxiv.org/abs/2006.03274}.

\bibitem[Hendrycks et~al.(2021)Hendrycks, Burns, Basart, Zou, Mazeika, Song, and Steinhardt]{hendryckstest2021}
Dan Hendrycks, Collin Burns, Steven Basart, Andy Zou, Mantas Mazeika, Dawn Song, and Jacob Steinhardt.
\newblock Measuring massive multitask language understanding.
\newblock \emph{Proceedings of the International Conference on Learning Representations (ICLR)}, 2021.

\bibitem[Kang et~al.(2024)Kang, Laroche, Yuan, Trischler, Liu, and Fu]{KangLYT0F24}
Jikun Kang, Romain Laroche, Xingdi Yuan, Adam Trischler, Xue Liu, and Jie Fu.
\newblock Think before you act: Decision transformers with working memory.
\newblock In \emph{{ICML}}. OpenReview.net, 2024.

\bibitem[Kaplan et~al.(2020)Kaplan, McCandlish, Henighan, Brown, Chess, Child, Gray, Radford, Wu, and Amodei]{kaplan2020scaling}
Jared Kaplan, Sam McCandlish, Tom Henighan, Tom~B Brown, Benjamin Chess, Rewon Child, Scott Gray, Alec Radford, Jeffrey Wu, and Dario Amodei.
\newblock Scaling laws for neural language models.
\newblock \emph{arXiv preprint arXiv:2001.08361}, 2020.

\bibitem[Kenton and Toutanova(2019)]{kenton2019bert}
Jacob Devlin Ming-Wei~Chang Kenton and Lee~Kristina Toutanova.
\newblock Bert: Pre-training of deep bidirectional transformers for language understanding.
\newblock In \emph{Proceedings of naacL-HLT}, volume~1, page~2. Minneapolis, Minnesota, 2019.

\bibitem[Ko et~al.(2024)Ko, Dai, Das, Kollias, Chaudhury, and Lozano]{ko2024memreasoner}
Ching-Yun Ko, Sihui Dai, Payel Das, Georgios Kollias, Subhajit Chaudhury, and Aurelie Lozano.
\newblock Memreasoner: A memory-augmented llm architecture for multi-hop reasoning.
\newblock In \emph{The First Workshop on System-2 Reasoning at Scale, NeurIPS'24}, 2024.

\bibitem[Kuratov et~al.(2024)Kuratov, Bulatov, Anokhin, Rodkin, Sorokin, Sorokin, and Burtsev]{kuratov2024babilong}
Yuri Kuratov, Aydar Bulatov, Petr Anokhin, Ivan Rodkin, Dmitry Sorokin, Artyom Sorokin, and Mikhail Burtsev.
\newblock Babilong: Testing the limits of llms with long context reasoning-in-a-haystack, 2024.

\bibitem[Lewis et~al.(2020)Lewis, Perez, Piktus, Petroni, Karpukhin, Goyal, K{\"{u}}ttler, Lewis, Yih, Rockt{\"{a}}schel, Riedel, and Kiela]{DBLP:journals/corr/abs-2005-11401}
Patrick S.~H. Lewis, Ethan Perez, Aleksandra Piktus, Fabio Petroni, Vladimir Karpukhin, Naman Goyal, Heinrich K{\"{u}}ttler, Mike Lewis, Wen{-}tau Yih, Tim Rockt{\"{a}}schel, Sebastian Riedel, and Douwe Kiela.
\newblock Retrieval-augmented generation for knowledge-intensive {NLP} tasks.
\newblock \emph{CoRR}, abs/2005.11401, 2020.
\newblock URL \url{https://arxiv.org/abs/2005.11401}.

\bibitem[Li et~al.(2023)Li, Cao, Kang, Yang, Chen, Jin, and Taylor]{li2023laffi}
Qianxi Li, Yingyue Cao, Jikun Kang, Tianpei Yang, Xi~Chen, Jun Jin, and Matthew~E Taylor.
\newblock Laffi: Leveraging hybrid natural language feedback for fine-tuning language models.
\newblock \emph{arXiv preprint arXiv:2401.00907}, 2023.

\bibitem[Liu and Lapata(2019)]{liu-lapata-2019-hierarchical}
Yang Liu and Mirella Lapata.
\newblock Hierarchical transformers for multi-document summarization.
\newblock In Anna Korhonen, David Traum, and Llu{\'i}s M{\`a}rquez, editors, \emph{Proceedings of the 57th Annual Meeting of the Association for Computational Linguistics}, pages 5070--5081, Florence, Italy, July 2019. Association for Computational Linguistics.
\newblock \doi{10.18653/v1/P19-1500}.
\newblock URL \url{https://aclanthology.org/P19-1500/}.

\bibitem[Loubna et~al.(2023)Loubna, Anton, and Elie]{huggingface_smollm}
Allal Loubna, Ben, Lozhkov Anton, and Bakouch Elie.
\newblock Small language models: Efficient, accessible, and effective.
\newblock \url{https://huggingface.co/blog/smollm}, 2023.
\newblock Accessed: 2025-01-16.

\bibitem[Mavi et~al.(2024)Mavi, Jangra, and Jatowt]{mavi2024multihopquestionanswering}
Vaibhav Mavi, Anubhav Jangra, and Adam Jatowt.
\newblock Multi-hop question answering, 2024.
\newblock URL \url{https://arxiv.org/abs/2204.09140}.

\bibitem[Penedo et~al.(2024)Penedo, Kydl{\'\i}{\v{c}}ek, Lozhkov, Mitchell, Raffel, Von~Werra, Wolf, et~al.]{penedo2024fineweb}
Guilherme Penedo, Hynek Kydl{\'\i}{\v{c}}ek, Anton Lozhkov, Margaret Mitchell, Colin Raffel, Leandro Von~Werra, Thomas Wolf, et~al.
\newblock The fineweb datasets: Decanting the web for the finest text data at scale.
\newblock \emph{arXiv preprint arXiv:2406.17557}, 2024.

\bibitem[Rodkin et~al.(2024)Rodkin, Kuratov, Bulatov, and Burtsev]{rodkin2024associativerecurrentmemorytransformer}
Ivan Rodkin, Yuri Kuratov, Aydar Bulatov, and Mikhail Burtsev.
\newblock Associative recurrent memory transformer, 2024.
\newblock URL \url{https://arxiv.org/abs/2407.04841}.

\bibitem[Weston et~al.(2016)Weston, Bordes, Chopra, and Mikolov]{westonBCM15}
Jason Weston, Antoine Bordes, Sumit Chopra, and Tom{\'{a}}s Mikolov.
\newblock Towards ai-complete question answering: {A} set of prerequisite toy tasks.
\newblock In \emph{{ICLR} (Poster)}, 2016.

\bibitem[Zaheer et~al.(2020)Zaheer, Guruganesh, Dubey, Ainslie, Alberti, Onta{\~{n}}{\'{o}}n, Pham, Ravula, Wang, Yang, and Ahmed]{DBLP:journals/corr/abs-2007-14062}
Manzil Zaheer, Guru Guruganesh, Avinava Dubey, Joshua Ainslie, Chris Alberti, Santiago Onta{\~{n}}{\'{o}}n, Philip Pham, Anirudh Ravula, Qifan Wang, Li~Yang, and Amr Ahmed.
\newblock Big bird: Transformers for longer sequences.
\newblock \emph{CoRR}, abs/2007.14062, 2020.
\newblock URL \url{https://arxiv.org/abs/2007.14062}.

\bibitem[Zhu et~al.(2020)Zhu, Xia, Wu, He, Qin, Zhou, Li, and Liu]{zhu2020incorporating}
Jinhua Zhu, Yingce Xia, Lijun Wu, Di~He, Tao Qin, Wengang Zhou, Houqiang Li, and Tie-Yan Liu.
\newblock Incorporating bert into neural machine translation.
\newblock \emph{arXiv preprint arXiv:2002.06823}, 2020.

\end{thebibliography}

\newpage
\appendix

\section{BABILong Dataset}
\label{sec:babilong_dataset}
This section provides an overview of the tasks in BABILong. 
Each task targets a specific aspect of language understanding and reasoning, 
forming a core benchmark for assessing model performance on retrieve factors from long context.

\begin{itemize}
    \item \textbf{Task 1: Single Supporting Fact} \\
    \textit{Goal}: Identify and use exactly one piece of relevant information from the text 
    to answer a question. \\
    \textit{Key Challenge}: Pinpointing the specific sentence or fact that directly yields 
    the correct answer.

    \item \textbf{Task 2: Two Supporting Facts} \\
    \textit{Goal}: Answer questions using two pieces of interconnected information. \\
    \textit{Key Challenge}: Linking separate facts and understanding how they combine 
    to produce the correct answer.

    \item \textbf{Task 3: Three Supporting Facts} \\
    \textit{Goal}: Extend the reasoning chain to three distinct pieces of information. \\
    \textit{Key Challenge}: Maintaining accuracy over longer inference chains 
    and managing multiple pieces of related text.

    \item \textbf{Task 4: Two Argument Relations} \\
    \textit{Goal}: Understand relationships involving two entities (arguments) 
    to answer questions. \\
    \textit{Key Challenge}: Correctly interpreting and manipulating relational information 
    (e.g., who gave what to whom) with two entities.

    \item \textbf{Task 5: Three Argument Relations} \\
    \textit{Goal}: Similar to Task 4 but introduces a third entity in the relationship. \\
    \textit{Key Challenge}: Tracking more complex interactions among three entities 
    while maintaining clarity and correctness.

    \item \textbf{Task 6: Yes/No Questions} \\
    \textit{Goal}: Provide binary (yes/no) answers based on the facts. \\
    \textit{Key Challenge}: Determining whether sufficient evidence exists in the text 
    to affirm or deny the query.

    \item \textbf{Task 7: Counting} \\
    \textit{Goal}: Count the number of times or entities that meet certain conditions. \\
    \textit{Key Challenge}: Performing numerical reasoning and accurately tracking 
    quantities within the text.

    \item \textbf{Task 8: Lists/Sets} \\
    \textit{Goal}: Gather all items satisfying specific criteria into a list or set. \\
    \textit{Key Challenge}: Aggregating multiple elements from different parts of the text 
    into a cohesive list/set.

    \item \textbf{Task 9: Simple Negation} \\
    \textit{Goal}: Handle statements containing negation. \\
    \textit{Key Challenge}: Understanding how negative statements (e.g., 
    ``John did not pick up the apple'') alter the truth value and impact the answer.

    \item \textbf{Task 10: Indefinite Knowledge} \\
    \textit{Goal}: Work with statements that contain incomplete or uncertain information. \\
    \textit{Key Challenge}: Managing and expressing knowledge not explicitly stated 
    (e.g., ``Someone picked up the apple, but we don't know who'').
\end{itemize}

\section{BABILong Benchmark Results}
\label{sec:babilong_results}

In Table~\ref{tab:detail_babilong}, we present the whole expeirments of compared models on BABILong benchmark.

\begin{table*}[ht]
\centering
\caption{Detailed performance of BABILong benchmark}
\small
\label{tab:detail_babilong}
\begin{tabular}{ccccccccccc}
\hline
model                                  & qa1                      & qa2                     & qa3                     & qa4                      & qa5                      & qa6                      & qa7                     & qa8                      & qa9                      & qa10                     \\ \hline
\multicolumn{11}{c}{0K}                                                                                                                                                                                                                                                                                           \\
Llama-3.2-1.2B                         & 54.0                     & 25.0                    & 29.0                    & 62.0                     & 59.0                     & 49.0                     & 14.0                    & 52.0                     & 41.0                     & 22.0                     \\
Llama-3.2-3.2B                         & 62.0                     & 37.0                    & 29.0                    & 64.0                     & 82.0                     & 53.0                     & 25.0                    & 53.0                     & 65.0                     & 56.0                     \\
vanilla-Llama-1.7B                     & 86.0                     & 57.0                    & 46.0                    & 59.0                     & 85.0                     & 83.0                     & 95.0                    & 79.0                     & 83.0                     & 77.0                     \\
RMT-1.7B                               & 85.0                     & 49.0                    & 49.0                    & 81.0                     & 95.0                     & 84.0                     & 82.0                    & 78.0                     & 85.0                     & 76.0                     \\
LM2-1.7B                               & 99.0                     & 89.0                    & 70.0                    & 88.0                     & 98.0                     & 95.0                     & 96.0                    & 97.0                     & 99.0                     & 94.0                     \\
\multicolumn{11}{c}{1K}                                                                                                                                                                                                                                                                                           \\
Llama-3.2-1.2B                         & 48.0                     & 22.0                    & 24.0                    & 55.0                     & 69.0                     & 49.0                     & 9.0                     & 31.0                     & 55.0                     & 33.0                     \\
Llama-3.2-1.2B-RAG                     & 51.0                     & 14.0                    & 19.0                    & 59.0                     & 80.0                     & 49.0                     & 10.0                    & 38.0                     & 40.0                     & 46.0                     \\
vanilla-Llama-1.7B                     & 31.0                     & 21.0                    & 44.0                    & 43.0                     & 71.0                     & 60.0                     & 71.0                    & 40.0                     & 67.0                     & 58.0                     \\
RMT-1.7B                               & 35.0                     & 26.0                    & 29.0                    & 33.0                     & 61.0                     & 50.0                     & 83.0                    & 41.0                     & 68.0                     & 53.0                     \\
LM2-1.7B                               & 85.0                     & 59.0                    & 72.0                    & 68.0                     & 91.0                     & 84.0                     & 96.0                    & 69.0                     & 82.0                     & 77.0                     \\
\multicolumn{11}{c}{2K}                                                                                                                                                                                                                                                                                           \\
Llama-3.2-1.2B                         & 44.0                     & 18.0                    & 19.0                    & 50.0                     & 64.0                     & 52.0                     & 18.0                    & 24.0                     & 55.0                     & 42.0                     \\
Llama-3.2-1.2B-RAG                     & 52.0                     & 11.0                    & 12.0                    & 49.0                     & 75.0                     & 48.0                     & 5.0                     & 33.0                     & 50.0                     & 43.0                     \\
LM2-1.7B                               & 58.0                     & 43.0                    & 64.0                    & 43.0                     & 87.0                     & 73.0                     & 93.0                    & 53.0                     & 75.0                     & 69.0                     \\
RMT-1.7B                               & 44.0                     & 21.0                    & 43.0                    & 41.0                     & 79.0                     & 47.0                     & 78.0                    & 41.0                     & 69.0                     & 51.0                     \\
vanilla-Llama-1.7B                     & 25.0                     & 22.0                    & 37.0                    & 34.0                     & 58.0                     & 60.0                     & 65.0                    & 38.0                     & 66.0                     & 58.0                     \\
\multicolumn{11}{c}{4K}                                                                                                                                                                                                                                                                                           \\
Llama-3.2-1.2B                         & 37.0                     & 16.0                    & 25.0                    & 56.0                     & 56.0                     & 50.0                     & 14.0                    & 27.0                     & 55.0                     & 32.0                     \\
Llama-3.2-1.2B-RAG                     & 47.0                     & 3.0                     & 16.0                    & 58.0                     & 68.0                     & 58.0                     & 3.0                     & 36.0                     & 45.0                     & 39.0                     \\
LM2-1.7B                               & 46.0                     & 37.0                    & 48.0                    & 34.0                     & 78.0                     & 66.0                     & 93.0                    & 45.0                     & 62.0                     & 50.0                     \\
RMT-1.7B                               & 24.0                     & 20.0                    & 22.0                    & 24.0                     & 28.0                     & 46.0                     & 75.0                    & 35.0                     & 65.0                     & 45.0                     \\
vanilla-Llama-1.7B                     & 21.0                     & 18.0                    & 38.0                    & 28.0                     & 55.0                     & 61.0                     & 64.0                    & 35.0                     & 49.0                     & 53.0                     \\
\multicolumn{11}{c}{8K}                                                                                                                                                                                                                                                                                           \\
Llama-3.2-1.2B                         & 26.0                     & 11.0                    & 24.0                    & 40.0                     & 52.0                     & 44.0                     & 25.0                    & 19.0                     & 44.0                     & 40.0                     \\
Llama-3.2-1.2B-RAG                     & 36.0                     & 1.0                     & 5.0                     & 57.0                     & 72.0                     & 49.0                     & 8.0                     & 28.0                     & 44.0                     & 35.0                     \\
LM2-1.7B                               & 34.0                     & 12.0                    & 31.0                    & 26.0                     & 63.0                     & 53.0                     & 95.0                    & 40.0                     & 57.0                     & 49.0                     \\
RMT-1.7B                               & 14.0                     & 15.0                    & 25.0                    & 28.0                     & 25.0                     & 47.0                     & 74.0                    & 38.0                     & 65.0                     & 46.0                     \\
vanilla-Llama-1.7B                     & 17.0                     & 19.0                    & 26.0                    & 20.0                     & 41.0                     & 51.0                     & 60.0                    & 37.0                     & 42.0                     & 45.0                     \\
\multicolumn{11}{c}{16K}                                                                                                                                                                                                                                                                                          \\
Llama-3.2-1.2B                         & 24.0                     & 6.0                     & 19.0                    & 33.0                     & 46.0                     & 55.0                     & 20.0                    & 13.0                     & 47.0                     & 48.0                     \\
Llama-3.2-1.2B-RAG                     & 26.0                     & 2.0                     & 9.0                     & 59.0                     & 76.0                     & 45.0                     & 5.0                     & 29.0                     & 52.0                     & 36.0                     \\
LM2-1.7B                               & 23.0                     & 17.0                    & 28.0                    & 28.0                     & 39.0                     & 44.0                     & 93.0                    & 38.0                     & 48.0                     & 42.0                     \\
RMT-1.7B                               & 23.0                     & 9.0                     & 18.0                    & 23.0                     & 19.0                     & 47.0                     & 75.0                    & 33.0                     & 62.0                     & 42.0                     \\
vanilla-Llama-1.7B                     & 10.0                     & 11.0                    & 21.0                    & 11.0                     & 37.0                     & 59.0                     & 61.0                    & 34.0                     & 46.0                     & 46.0                     \\
\multicolumn{11}{c}{32K}                                                                                                                                                                                                                                                                                          \\
Llama-3.2-1.2B                         & 15.0                     & 7.0                     & 15.0                    & 24.0                     & 46.0                     & 54.0                     & 23.0                    & 13.0                     & 53.0                     & 46.0                     \\
Llama-3.2-1.2B-RAG                     & 28.0                     & 1.0                     & 2.0                     & 51.0                     & 74.0                     & 51.0                     & 2.0                     & 19.0                     & 41.0                     & 32.0                     \\
LM2-1.7B                               & 19.0                     & 13.0                    & 20.0                    & 23.0                     & 31.0                     & 50.0                     & 92.0                    & 35.0                     & 59.0                     & 39.0                     \\
RMT-1.7B                               & 12.0                     & 16.0                    & 20.0                    & 18.0                     & 22.0                     & 46.0                     & 74.0                    & 34.0                     & 62.0                     & 43.0                     \\
vanilla-Llama-1.7B                     & 10.0                     & 17.0                    & 24.0                    & 13.0                     & 30.0                     & 54.0                     & 71.0                    & 33.0                     & 39.0                     & 53.0                     \\
\multicolumn{11}{c}{64K}                                                                                                                                                                                                                                                                                          \\
Llama-3.2-1.2B                         & 11.0                     & 8.0                     & 13.0                    & 12.0                     & 42.0                     & 43.0                     & 17.0                    & 6.0                      & 48.0                     & 33.0                     \\
Llama-3.2-1.2B-RAG                     & 27.0                     & 0.0                     & 4.0                     & 56.0                     & 66.0                     & 54.0                     & 4.0                     & 15.0                     & 48.0                     & 44.0                     \\
LM2-1.7B                               & 19.0                     & 18.0                    & 19.0                    & 19.0                     & 22.0                     & 42.0                     & 91.0                    & 35.0                     & 51.0                     & 38.0                     \\
RMT-1.7B                               & 21.0                     & 18.0                    & 19.0                    & 21.0                     & 15.0                     & 48.0                     & 70.0                    & 33.0                     & 61.0                     & 41.0                     \\
vanilla-Llama-1.7B                     & 8.0                      & 13.0                    & 14.0                    & 14.0                     & 16.0                     & 52.0                     & 60.0                    & 30.0                     & 41.0                     & 41.0                     \\
\multicolumn{11}{c}{128K}                                                                                                                                                                                                                                                                                         \\
\multicolumn{1}{l}{Llama-3.2-1.2B-RAG} & \multicolumn{1}{l}{17.0} & \multicolumn{1}{l}{0.0} & \multicolumn{1}{l}{3.0} & \multicolumn{1}{l}{51.0} & \multicolumn{1}{l}{73.0} & \multicolumn{1}{l}{49.0} & \multicolumn{1}{l}{5.0} & \multicolumn{1}{l}{11.0} & \multicolumn{1}{l}{46.0} & \multicolumn{1}{l}{39.0} \\
LM2-1.7B                               & 15.0                     & 16.0                    & 12.0                    & 19.0                     & 23.0                     & 48.0                     & 91.0                    & 34.0                     & 54.0                     & 38.0                     \\
RMT-1.7B                               & 17.0                     & 13.0                    & 20.0                    & 21.0                     & 18.0                     & 47.0                     & 72.0                    & 35.0                     & 64.0                     & 42.0                     \\
vanilla-Llama-1.7B                     & 7.0                      & 14.0                    & 19.0                    & 12.0                     & 13.0                     & 52.0                     & 63.0                    & 28.0                     & 46.0                     & 42.0                     \\ \hline
\end{tabular}
\end{table*}

\end{document}